%% file: main.tex
\documentclass[twocolumn]{article}
\usepackage{spconf,amsmath,graphicx}
\usepackage{amssymb}
\usepackage{float}
\usepackage{tabularray}
\usepackage{lipsum}
\usepackage{multirow, booktabs}
\usepackage{tikz}
\def\checkmark{\tikz\fill[scale=0.4](0,.35) -- (.25,0) -- (1,.7) -- (.25,.15) -- cycle;}

\title{A Density-Guided Temporal Attention Transformer for Indiscernible Object Counting in Underwater Videos}

\name{Cheng-Yen Yang$^{1}$\sthanks{Contact author: cycyang@uw.edu. $^{**}$ This work was funded by Fisheries Southeast Fisheries Science Center of National Oceanic and Atmospheric Administration (US).}, Hsiang-Wei Huang$^{1}$, Zhongyu Jiang$^{1}$, Hao Wang$^{1}$, Farron Wallace$^{2}$, Jenq-Neng Hwang$^{1}$}

\address{$^{1}$ Department of Electrical $\&$ Computer Engineering, University of Washington, USA\\
$^{2}$ National Oceanic and Atmospheric Administration (NOAA), USA}

\begin{document}

\maketitle

\begin{abstract}
Dense object counting or crowd counting has come a long way thanks to the recent development in the vision community. However, indiscernible object counting, which aims to count the number of targets that are blended with respect to their surroundings, has been a challenge. Image-based object counting datasets have been the mainstream of the current publicly available datasets. Therefore, we propose a large-scale dataset called YoutubeFish-35, which contains a total of 35 sequences of high-definition videos with high frame-per-second and more than 150,000 annotated center points across a selected variety of scenes.  For bench-marking purposes, we select three mainstream methods for dense object counting and carefully evaluate them on the newly collected dataset. We propose TransVidCount, a new strong baseline that combines density and regression branches along the temporal domain in a unified framework and can effectively tackle indiscernible object counting with state-of-the-art performance on YoutubeFish-35 dataset.
\end{abstract}


\input{table/dataset_overview}

\section{Introduction}
\label{sec:intro}

Object counting is a critical task in computer vision, involving the counting of specific object instances in images or videos. While it may seem similar to object detection, object counting presents unique challenges beyond traditional detection methods. These challenges include dealing with overlapping instances, partial occlusion, varying object sizes, crowded scenes, density estimation, specialized validation metrics, and the need for tailored model design. There is a growing interest in dense object-related tasks \cite{sun2023iocfish}, particularly for crowded scenes, driven by various real-world applications such as event monitoring \cite{chan2008ucsd, chen2012mall}, surveillance security \cite{schroder2018crowdflow, fang2019fdst}, commercial fishery management \cite{zhang2020fish}, and wildlife preservation \cite{arteta2016countinginthewild}. These applications all require precise counting of densely clustered objects in crowded environments.

Recently, the focus on object counting within indiscernible or camouflaged scenes has intensified due to the unexplored nature of this field. Studies such as Camouflaged Object Detection (COD) or Camouflaged Segmentation (CIS), along with Indiscernible Object Counting (IOC) have delved into this area. As defined in \cite{sun2023iocfish}, an indiscernible scene pertains to situations where foreground objects closely resemble the background in terms of patterns, colors, or textures.

\begin{figure}[tb]
\centering
\caption{Visualizations of the dataset: YoutubeFish-35, the first video-based dataset for indiscernible object counting with point-level annotations.}
\includegraphics[width=0.95\linewidth]{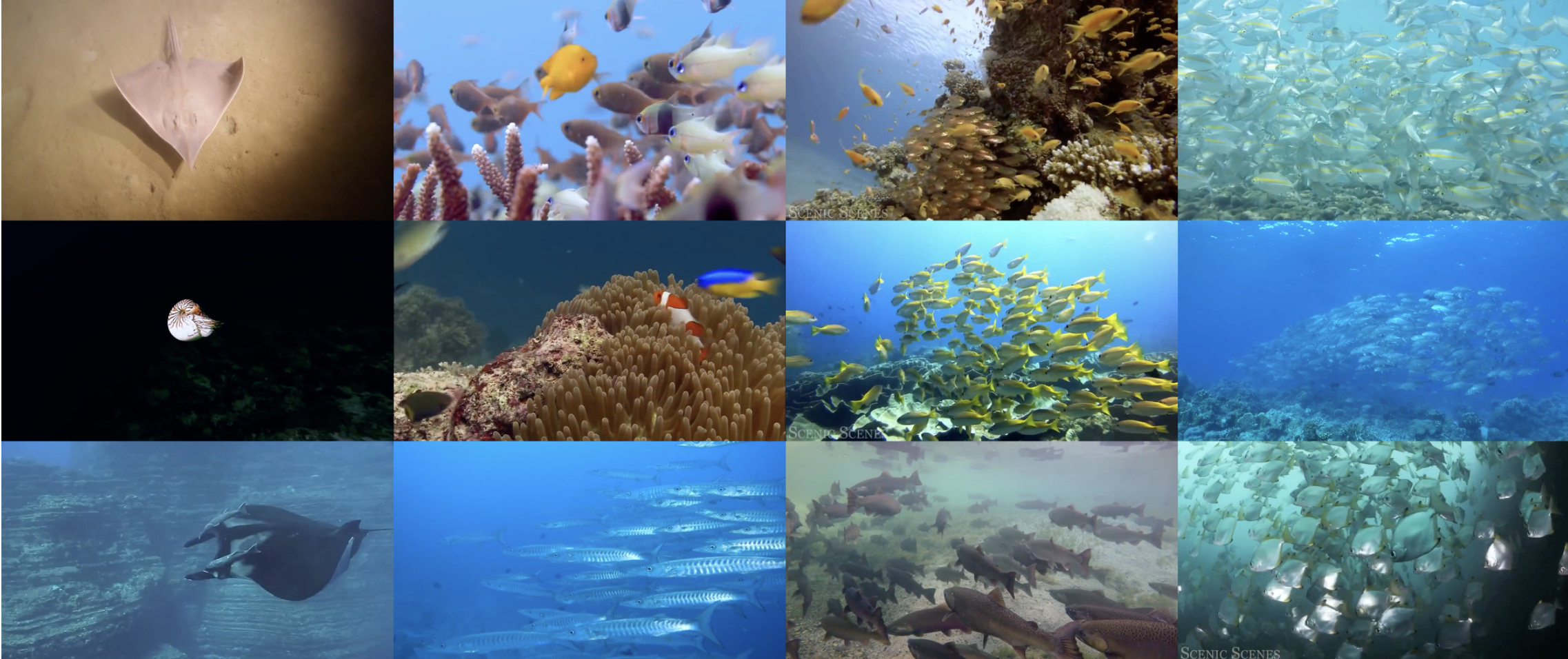}
\label{fig:one}
\end{figure}

Although several datasets with instance-level or point-level annotations can facilitate research in Indiscernible Object Counting (IOC), a dedicated video-based dataset capturing indiscernible or camouflaged scenes would be particularly beneficial to IOC research. Such a dataset could provide a better understanding of the challenges associated with object counting in these complex scenarios, enabling the development and evaluation of more robust IOC techniques.

To facilitate the research on Indiscernible Object Counting (IOC) and align it more closely with real-world applications, we introduced a video-based dataset, YoutubeFish-35. This dataset offers high-definition videos that capture scenes where objects are challenging to discern due to high blending into their surroundings. By providing a diverse range of scenarios and challenges, YoutubeFish-35 propels IOC research toward solutions that can effectively address the intricacies of object counting in complex and realistic scenarios.

\section{Related Works}
\label{sec:related}

\vspace{-.5em}

\noindent \textbf{Image-based Object Counting.} Image-based object counting has extensively employed CNNs and can be broadly categorized into three following directions. Detection-based approaches \cite{topkaya2014counting} identify individual instances through object detection models and subsequently count instances. Regression-based methods directly predict object counts, utilizing handcrafted features or deep learning architectures to learn the mapping between image features and counts. The latest work like CLTR \cite{liang2022cltr} proposed an end-to-end transformer-based network to count and localize the crowd. Density-based methods \cite{lin2022man, sun2023iocfish} focus on estimating object density, leading to the prediction of density maps for count integration. IOCFormer, on the other hand, adds the density branch on top of \cite{liang2022cltr} to achieve satisfying results on the IOC task. These directions represent diverse strategies to address the multifaceted challenge of crowd counting, each offering unique advantages and limitations.

\noindent \textbf{Video-based Object Counting.} Video-based methods have fewer options than single-image techniques despite videos being more practical for real-world applications. ConvLSTM \cite{xiong2017convlstm} utilizes a bidirectional convolutional LSTM to capture temporal information. MOPN \cite{hossain2020mopn} employs a pyramid-like optical flow structure to decode motion information for count estimation while Locality-constrained Spatial Transformer (LSTN) \cite{fang2019lstn} uses a spatial transformer network to predict the density map of consecutive frames.  The Spatial-Temporal Graph Network (STGN) \cite{wu2022stgn} efficiently and accurately counts crowds in videos by learning relations between pixels and patches in local spatial-temporal domains. Nevertheless, all the aforementioned and existing endeavors predominantly revolve around tallying human crowds within videos, which could potentially disregard the imperative task of counting objects within indiscernible scenes. 

\begin{figure*}[h]
\caption{The overview of our proposed method: TransVidCount.}
\centering
\includegraphics[width=\textwidth]{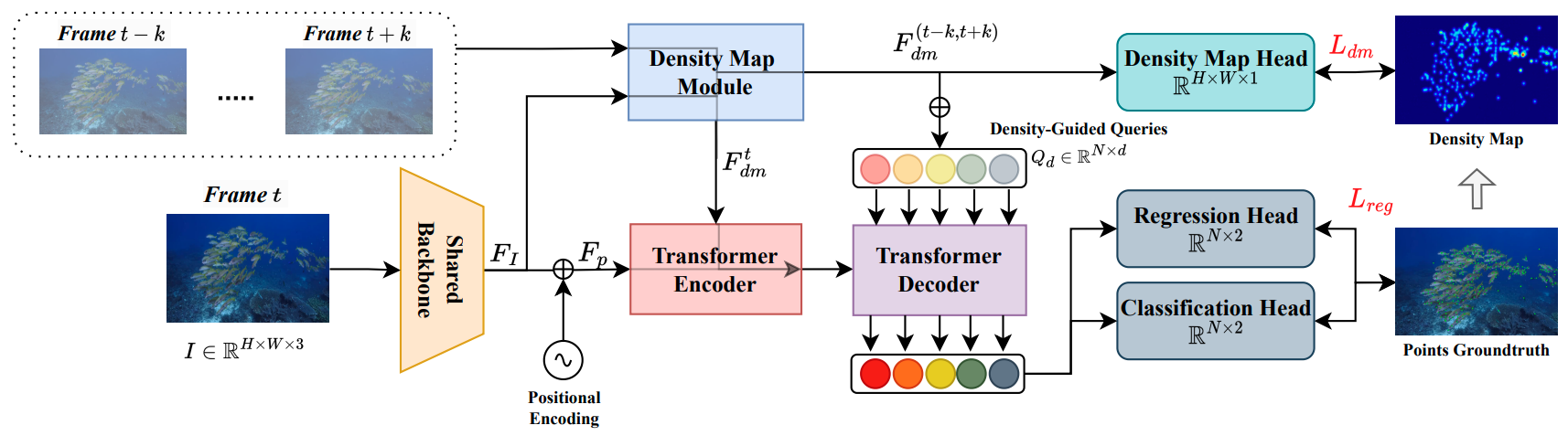}
\end{figure*}

\vspace{-.5em}

\section{Proposed Dataset: YoutubeFish-35}
\label{sec:dataset}


\noindent \textbf{Data Collection.} We were inspired by the IOCfish5K dataset \cite{sun2023iocfish} and decided to create a new dataset. Our focus was on selecting high-definition free-view video sequences that contain temporal information. This allowed us to extend object counting to video-based methods, which are often more applicable to real-world situations. Our dataset follows the definition of indiscernible objects for underwater scenes. In the first data collection stage, we manually selected and downloaded 35 different high-quality videos from YouTube. The video sequences are mostly recorded between 25 to 30 fps in 1080 HD resolution. 

\noindent \textbf{Annotation Pipeline.} We selected 35 individual 150-frame segments which we think have the most diverse environments (e.g., ocean, lake, river), lighting conditions (e.g., muddy, dark with light), and movement (e.g., sudden movement, swimming in school) as in Fig.~\ref{fig:one}. We followed the definition and annotation principle of IOCfish5K, in which the goal was to annotate each animal with a point at the center of its visible part using a modified version of the labelme software with custom features. This step takes around 250 human hours.

\vspace{-1em}

\section{Proposed Method}
\label{sec:method}

\vspace{-.5em}

The overall architecture of our proposed method is an end-to-end network that takes consecutive frames as input, which is shown in Fig. 2. There are three main components of the network: (1) a shared CNN-based backbone, (2) a density map module, (3) an encoder-decoder transformer \cite{carion2020detr, meng2021cdetr}.

\vspace{-1em}

\subsection{Density Map Module}

Employing pseudo density maps to semi-supervise predicted counts from generated density maps holds significant potential. This approach elegantly encodes the region's crowd density and provides valuable localization information. A sequence of images will be fed into a shared backbone to generate image features ${F}_{I}^{t}$ followed by a series of convolution blocks to generate intermediate density features  ${F}^{t}_{dm}\in\mathbb{R}^{h'\times w'\times d'}$.  Then, the density map head with $1\times1$ convolution layer will output the final density map prediction $D \in \mathbb{R}^{h\times w\times 1}$. The Euclidean distance is used to measure the difference between the estimated density map ${D}^{i}_{dm}$  and ground truth. The loss is defined as:

\begin{equation}
    L_{dm} = \frac{1}{N} \sum^{N}_{i=1} || D^{i}_{dm} - D^{i}_{pdm} ||^2_{2}
\end{equation}

\subsection{Temporal Density-Guided Transformer}

\noindent \textbf{Encoder.} Inspired by the density-enhanced transformer encoder (DETE) proposed in \cite{sun2023iocfish}, we further implement and integrate a temporal-attention mechanism to allow the encoder part to capture better the temporal information to boost the representations.  Similarly, the encoder is made up of $L$ layers of Multi-head Self-Attention (MSA) blocks, where:

\begin{equation}
    F'_{l-1} = MSA(LN(F_{l-1} + TA (F_{dm}))) 
\end{equation}
\begin{equation}
    F_{l} = F'_{l-1} + FC \big( LN(F'_{l}) \big)
\end{equation}

\noindent as $LN(\cdot)$ stands for layer normalization, $TA(\cdot)$ stands for temporal attention layer and $FC(\cdot)$ stands for feed-forward layers while $l=1,2,...,L$. And the $F_{dm}$ denotes the sets of density maps that are properly reshaped by a series of convolutional layers to match the dimension of input or output of each layer of the transformer encoder blocks. Each independent self-attention module consists of three inputs: query, key, and value, which are computed from $F_{l-1}$:

\begin{equation}
    SA(Q,K,V) = softmax(\frac{QK^T}{\sqrt{D}})V,
\end{equation}
as the standard implementation in most transformer frameworks \cite{carion2020detr, meng2021cdetr, liang2022cltr, sun2023iocfish}, where $Q$, $K$, and $V$ represents queries, keys and values. $D$ stands for the dimension of features.

\noindent \textbf{Decoder.} The original implementation of the decoder of the regression branch \cite{liang2022cltr} follow CDETR \cite{carion2020detr, meng2021cdetr} as a trainable query $Q$ was used as input for the decoder when cross-attention is performed between the queries and the input feature map, the temporal information of density maps are embedded to the queries $Q$ to make point predictions. 

The current method for training queries relies on examining the entire patch to locate each target. Therefore, we argue that by incorporating the density map information, the queries can be properly guided toward better understanding and interpretation when trying to localize and classify whether it is an object or not. The density-guided queries are built upon the conditional cross-attention:

\begin{equation}
    Q_d = W_Q + F(conv(TA(F_{dm}))),
\end{equation}

\noindent The decoder output will be sent to the regression head and classification head to generate coordinates along with confidence score predictions.

\subsection{Loss Function}

Following the acquisition of one-to-one matching results and obtaining the generated density map, we proceed with the calculation of the loss for subsequent backpropagation. Our approach involves direct point predictions, with the resultant loss encompassing all three of the density map generation, point regression, and classification components. 

Finally, the localization loss $L_{loc}$ computes the $L_1$ distance between predicted points and ground-truth coordinates while the $L_{cls}$ is the focal loss \cite{lin2017focal} for target or background. The end-to-end loss can be written in the form:

\begin{equation}
    L_{total} = \lambda_{reg}\cdot(L_{cls} + L_{loc}) + \lambda_{dm} \cdot L_{dm}
\end{equation}

\noindent where $\lambda_{reg}=1$ and $\lambda_{dm}=0.25$. The entire model is jointly trained, and during inference, we take the point predictions as the result.

\begin{figure*}[h]
\caption{Visualization of the count estimation of CLTR \cite{liang2022cltr}, IOCFormer \cite{sun2023iocfish} and TransVidCount. The first row contains the input images and the corresponding ground-truth counts, while the latter rows represent the predicted counts and their coordinates.}
\centering
\includegraphics[width=0.92\textwidth]{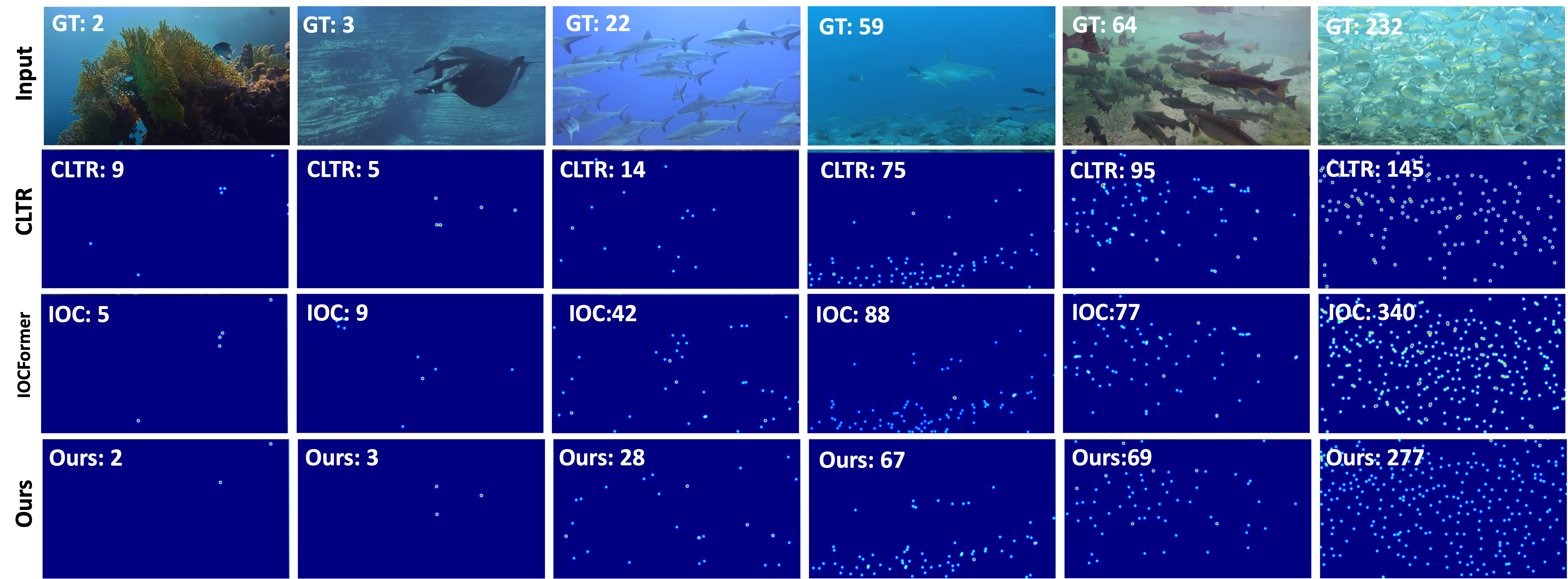}
\end{figure*}

\section{Experimental Results}
\label{sec:exp}

\subsection{Experiment Settings}

\noindent\textbf{Implementation Details.} There are a total of 25 sequences in training set, $3$ sequences in validation set, and $7$ sequences in testing set. We conduct our experiments using PyTorch and $4$ Nvidia Tesla V100 for training. For a fair comparison, we adapt the official implementation of MAN \cite{lin2022man}, CLTR \cite{liang2022cltr}, and IOCFormer \cite{sun2023iocfish} and used similar hyper-parameters during training. All methods are trained for 750 epochs with the exact same learning rate setting. During the inference phase, we divide the images into patches of identical dimensions as those employed during the training process. We apply a threshold of 0.3 to eliminate predictions related to the background.

\noindent \textbf{Evaluation Metrics.} To assess the performance of the proposed method, we calculate the Mean Absolute Error (MAE), Mean Square Error (MSE), and Mean Normalized Absolute Error (NAE) by comparing predicted counts with ground-truth counts across all images following \cite{lin2022man, liang2022cltr, sun2023iocfish}. No additional post-processing are being made to the predictions of all the compared methods.

\subsection{Quantitative Results}

\noindent \textbf{Results on YoutubeFish-35.} We compare our proposed method with some of the latest crowd counting and indiscernible object counting techniques, as showcased in the results of the testing split of our newly introduced dataset in Table 2. Our approach leverages temporal information and demonstrates its ability to provide more accurate counts in dynamic scenarios, where object densities change over time. Visualizations of the predicted counts and coordinates are presented in Fig. 3 as our TransVidCount can accurately locate some of the highly-occluded targets in comparison to other image-based methods. These enhanced queries improve our single-frame method by an 11\% margin and video-based method by a 15\% margin over the existing SOTA.

\noindent \textbf{Ablation Experiment.} We extend our analysis to compare the various operations employed during the aggregation of density-guided queries, as detailed in Table 3. The outcomes of this comparative evaluation serve to affirm the efficacy of our approach utilizing density-guided queries in augmenting the features. This is accomplished by leveraging the information derived from the consecutive density maps, thus providing substantial validation for the utility of our proposed method in enhancing feature extraction and representation.

\begin{table}[t]
\centering
\small
\caption{Comparison of the evaluation metrics (MAE, MSE, and NAE) on the YoutubeFish-35 testing split.}
\begin{tabular}{l|c|c|c}
\hline
Method              & MAE $\downarrow$ & MSE $\downarrow$ & NAE $\downarrow$ \\ \hline
MAN (CVPR'22)       & 35.952    & 44.533    &  1.056   \\
CLTR (ECCV'22)      & 21.068    & 36.630    & 0.597    \\
IOCFormer (CVPR'23) & 16.194    & 22.005    &  0.493   \\ \hline
TransVidCount (frame=1)       & 14.372    & 19.194    &  0.514   \\
TransVidCount (frame=5)      & \textbf{13.714}    & \textbf{17.909}    &  \textbf{0.394}   \\ \hline
\end{tabular}
\end{table}

\begin{table}[t]
\centering
\small
\caption{Ablation study of the density-guided methods.}
\begin{tabular}{l|c|c|c|c}
\hline
Method              & Operation &  MAE $\downarrow$ & MSE $\downarrow$ & NAE $\downarrow$ \\ \hline
       & -    & 16.194    & 22.005    &  0.493    \\
Baseline       & Add.    & \textbf{13.481}    & 18.359    &  0.472   \\
       & Concat.    & 13.714    & \textbf{17.909}    &  \textbf{0.394}   \\ \hline
\end{tabular}
\end{table}


\noindent \textbf{Limitations.} We want to further discuss some limitations of the IOC task and our framework. First, larger objects (or objects that are closer to the camera) often resulted in duplicated prediction due to a fixed cropping size being adapted in such object counting frameworks. This can either be addressed by resizing the image or equipping post-processing with detection results. We would also like to mention that the latency ($5$-frames) is roughly 30$\%$ higher than the baseline model ($1$-frame), which might be some trade-off that we need to evaluate when deploying to real-world applications.

\section{Conclusion}

In conclusion, our research presents a new video-based dataset and a novel approach for counting indiscernible objects in underwater videos, addressing the challenge of counting obscured and hard-to-discern objects and providing a benchmark for future studies. TransVidCount, with its attention mechanisms and temporal modeling, outperforms traditional methods and other deep learning architectures, showcasing its effectiveness in improving count accuracies.

\newpage
\newpage

\bibliographystyle{IEEEbib}
\bibliography{strings,refs}

\end{document}

%% file: table/dataset_overview.tex
\begin{table*}
\centering
\footnotesize
\caption{Overview of the publicly available dense object (or crowd) counting datasets.}
\label{table:dataset}
\begin{tblr}{
  stretch = 0,
  row{2} = {c},
  column{2} = {c},
  column{3} = {c},
  column{4} = {c},
  column{5} = {c},
  column{6} = {c},
  cell{1}{1} = {r=2}{},
  cell{1}{2} = {r=2}{},
  cell{1}{3} = {r=2}{},
  cell{1}{4} = {r=2}{},
  cell{1}{5} = {r=2}{},
  cell{1}{6} = {r=2}{},
  cell{1}{7} = {c=4}{c},
  cell{3}{3} = {c},
  cell{3}{7} = {c},
  cell{3}{8} = {c},
  cell{3}{9} = {c},
  cell{3}{10} = {c},
  cell{4}{3} = {c},
  cell{4}{7} = {c},
  cell{4}{8} = {c},
  cell{4}{9} = {c},
  cell{4}{10} = {c},
  cell{5}{3} = {c},
  cell{5}{7} = {c},
  cell{5}{8} = {c},
  cell{5}{9} = {c},
  cell{5}{10} = {c},
  cell{6}{3} = {c},
  cell{6}{7} = {c},
  cell{6}{8} = {c},
  cell{6}{9} = {c},
  cell{6}{10} = {c},
  cell{7}{3} = {c},
  cell{7}{7} = {c},
  cell{7}{8} = {c},
  cell{7}{9} = {c},
  cell{7}{10} = {c},
  cell{8}{3} = {c},
  cell{8}{7} = {c},
  cell{8}{8} = {c},
  cell{8}{9} = {c},
  cell{8}{10} = {c},
  cell{9}{3} = {c},
  cell{9}{7} = {c},
  cell{9}{8} = {c},
  cell{9}{9} = {c},
  cell{9}{10} = {c},
  vline{2-7} = {1}{},
  vline{10} = {2}{},
  vline{2-7,10} = {2-9}{},
  hline{1,3,6,9-10} = {-}{},
  hline{2} = {7-10}{},
}
Datasets              & Type  & IOC & {Camera\\View} & \# of Seq & {Average\\Resolution} & \# of Image &             &       &       \\
                      &       &                        &             &           &                       & $<50$          & $51\sim200$ & $>200$   & Total \\
NC4K \cite{lv2021nc4k}                  & Image &   \checkmark                     &   -          & -         & 709$\times$530               & 4,121       & 0           & 0     & 4,121 \\
COD \cite{fan2021cod}                  & Image &  \checkmark                      &   -          & -         & 964$\times$737               & 5,066       & 0           & 0     & 5,066 \\
IOCfish5K \cite{sun2023iocfish}            & Image &   \checkmark                     &   -          & -         & 1920$\times$1080               & 2,663       & 1,957       & 1,017 & 5,637 \\
UCSD \cite{chan2008ucsd}                 & Video &     -                   & Fixed       & 1         & 238$\times$158               & 2,000       & 0           & 0     & 2,000 \\
Mall \cite{chen2012mall}                 & Video &     -                   & Fixed       & 1         & 640$\times$480               & 2,000       & 0           & 0     & 2,000 \\
CrowdFlow \cite{schroder2018crowdflow}            & Video &   -                     & Moving      & 5         & 1280$\times$720              & 0           & 0           & 3,200 & 3,200 \\
YoutubeFish-35 (Ours) & Video &      $\checkmark$                 & Moving      & 35        & 1920$\times$1080             & 4,235       & 715         & 300   & 5,250 \\
\end{tblr}
\end{table*}